\documentclass[11pt,letterpaper]{article}
\usepackage{emnlp2016}
\usepackage{times}
\usepackage{latexsym}

\usepackage{epsfig}
\usepackage{graphicx}
\usepackage{amsmath}
\usepackage{amssymb}
\usepackage{amsfonts}
\usepackage{color}

\usepackage{algorithm}
\usepackage{multirow}
\usepackage{rotating}
\usepackage{booktabs}
\usepackage{slashbox} % Package to make a diagonal in a table entry
\usepackage{algpseudocode}
\usepackage[belowskip=-10pt,aboveskip=2pt,font=small]{caption}
\usepackage{subcaption}
\usepackage{soul,xcolor}
\usepackage{enumitem}
\usepackage[olditem,oldenum]{paralist}

\usepackage[normalem]{ulem}

\usepackage{mysymbols}
\usepackage{nicefrac}

\graphicspath{ {images/} }

\usepackage[pagebackref=true,breaklinks=true,letterpaper=true,colorlinks,bookmarks=false,citecolor=green,linkcolor=red]{hyperref}

\emnlpfinalcopy

\def\emnlppaperid{503}

\title{Towards Transparent AI Systems: \\ Interpreting Visual Question Answering Models}

\author{Yash Goyal, Akrit Mohapatra, Devi Parikh, Dhruv Batra \\
        Virginia Tech \\ 
        {\tt\small \{ygoyal, akrit, parikh, dbatra\}@vt.edu}}
\date{}

\begin{document}

\maketitle

\begin{abstract}

Deep neural networks have shown striking progress and obtained state-of-the-art results in many AI research fields in the recent years.
However, it is often unsatisfying to not know why they predict what they do. 
In this paper, we address the problem of interpreting Visual Question Answering (VQA) models. 
Specifically, we are interested in finding what part of the input (pixels in images or words in questions) the VQA model focuses on while answering the question.
To tackle this problem, we use two visualization techniques -- guided backpropagation and occlusion -- to find important words in the question and important regions in the image.
We then present qualitative and quantitative analyses of these importance maps.
We found that even without explicit attention mechanisms, VQA models may sometimes be implicitly attending to relevant regions in the image, and often to appropriate words in the question.
\end{abstract}

\section{Introduction}
\label{sec:intro}

We are witnessing an excitement in the research community and frenzy in media regarding advances in AI. 
% Bring this back in camera-ready
Fueled by a combination of massive datasets and advances in deep neural networks (DNNs), the community has made remarkable progress 
in the past few years 
on a variety of `low-level' AI tasks such as image classification \cite{GoogleNet}
%scene recognition \cite{zhou2014learning}, 
%object detection \cite{FasterRCNN}, 
%semantic segmentation \cite{hariharan2015hypercolumns} 
%in computer vision (CV), 
machine translation \cite{brea2011sequence,rnn_mt} 
%in natural language processing (NLP), 
and speech recognition \cite{hinton2012deep}. 
Neural networks are also demonstrating potential in `high-level' AI tasks such as
learning to play 
%Atari video games and 
Go \cite{masteringGo}, 
%image captioning \cite{captioning_google,captioning_msr}, 
answering reading comprehension questions by understanding short stories \cite{bordes2015large,weston_qa}, 
and even answering questions about images \cite{VQA,Ren_2015_NIPS,Malinowski_2015_ICCV}.

% \sout{
% Fueled by a combination of massive datasets and advances in deep neural networks (DNNs), the community has made remarkable progress 
% on a variety of AI tasks such as image classification \cite{GoogleNet},
% machine translation \cite{brea2011sequence,rnn_mt}, 
% speech recognition \cite{hinton2012deep},
% and even harder tasks such as
% learning to play Go \cite{mnih2013playing}, 
% answering reading comprehension questions by understanding short stories \cite{bordes2015large,weston_qa}, 
% and answering questions about images \cite{VQA,Ren_2015_NIPS,Malinowski_2015_ICCV}.}

Unfortunately, when today's machine perception and intelligent systems fail, they fail in a spectacularly disgraceful manner, without warning or explanation, leaving the user staring at an incoherent output, wondering why the system did what it did.

\begin{figure}[t]
%\vskip 0.2in
% \begin{center}
\centering
\includegraphics[width=\columnwidth]{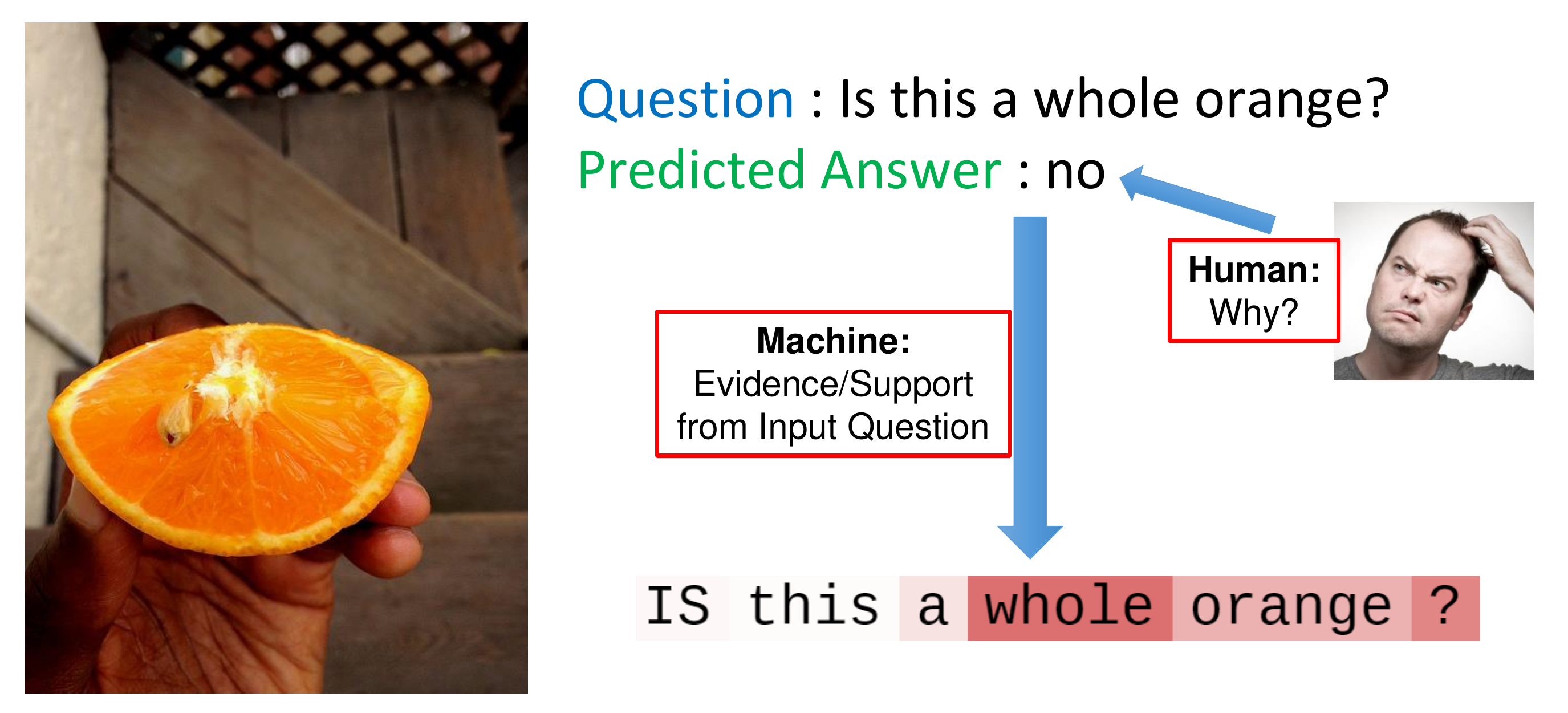}
% \vskip -0.15in
% \caption{An example illustrating the task of Visual Question Answering.
% The model predicts the answer ``no'' for the question ``Is this a whole orange?''.
% But, when the word ``whole'' is occluded in the question, the model's prediction changes to ``yes''.
% This demonstrates that the word ``whole'' is indeed important for the VQA model in predicting the answer.}
\caption{The goal of this work is to interpret Visual Question Answering (VQA) models. We are interested in answering the question-- \emph{why does a VQA model predict what it does?} Our approach here is to find the evidence in the test input on which the model focuses while answering the question. In this example, ``whole'' is the most important word in the question for the model while predicting the answer ``no''. 
%\comment{should include the corresponding image heat map as well?}
}
\label{fig:teaser}
% \end{center}
%\vspace{-10pt} 
\end{figure} 

In this work, we focus on 
Visual Question Answering, where given an image and a free-form natural language question about the image,
(e.g., ``What color are the girl's shoes?'', or ``Is the boy jumping?''),
the machine has to produce a natural language answer as its output
(e.g. ``blue'', or ``yes''). 
Specifically, we try to interpret a recent state-of-art VQA model \cite{Lu2015} trained on recently released VQA \cite{VQA} dataset.
This VQA model uses Convolutional Neural Network (CNN) based embedding of the image, Long-Short Term Memory (LSTM) based embedding of the question, combines these two embeddings, and uses a multi-layer perceptron as a classifier to predict a probability distribution over answers.

Sustained interactions\footnote{Demo available here: \url{http://cloudcv.org/vqa/} \cite{cloudcv}} with the system make it clear that the system has a non-trivial level of intelligence (e.g., it is able to recognize people, objects, \etc in the image). However, it is deeply unsatisfying to not know why the system predicts what it does (especially the glaring mistakes).
The root cause is \emph{lack of transparency or interpretability}. 
% Bring back the following in camera-ready
With a few rare exceptions, the emphasis in machine learning, computer vision, and AI communities today is on building systems with good predictive performance, not \emph{transparency}. As a result, the users of these intelligent systems perceive them as inscrutable black boxes that cannot be understood or trusted.

We are interested in the question of transparency -- 
\textit{why does a VQA system do what it does?} (See \figref{fig:teaser}).
Specifically, \emph{what evidence in the test input (image and question) supports a particular prediction?} In the context of VQA, this question can be expressed as two subproblems:
\begin{compactitem}
\item \textit{What words in the question does the model ``listen to'' in order to answer the question? }
\item \textit{What pixels in the image does the model ``look at'' while answering the question?}
\end{compactitem}
	
In this work, we use two visualization methods to tackle the above problems.
The first method (\secref{sec:guidedback}) uses guided backpropagation \cite{GuidedBackprop} to analyze important words in the question and important regions in the image. 
In the second method (\secref{sec:occlusion}), we occlude portions of input and observe the change in prediction probabilities of the model, to compute importance of question words and image regions.
In \secref{sec:results}, we present qualitative and quantitative analyses of these image/question `importance maps' -- question importance maps are analyzed using their Part-of-Speech (POS) tags; image importance maps are compared to `human attention maps' or maps showing where humans look for answering a question about the image \cite{human_attention}. 
We found that even without explicit attention mechanisms, VQA models may sometimes be implicitly attending to relevant regions in the image, and often to appropriate words in the question.
% Finally, we discuss in \secref{sec:failure_prediction} that these importance maps for questions and images contain cues which can be helpful for predicting the VQA model's failure in answering the question correctly.
\section{Related Work}
\label{sec:related_work}

Many gradient based methods \cite{Zeiler2014,SimonyanVZ13,GuidedBackprop} have been proposed in recent years in the field of computer vision to visualize deep convolutional neural networks. 
But most of them focused on the task of image classification on iconic images where the main object occupies most of the image. 
Our work differs in 2 ways -- 1) we also compute gradients \wrt the input question, and 2) we use guided backpropagation \cite{GuidedBackprop} for the task of VQA, where the model can look at different regions in the same image for different questions. As per our literature review, we are the first to study this problem for VQA.

Our occlusion experiment is inspired by \cite{Zeiler2014} who mask small regions in the image with a gray patch, and observe the output of an image classification model.
We evaluate if the model looks at the same regions in the image as humans do, while answering a question about the image.

A few recent works~\cite{sameersingh,Baehrens,LiuViz}
have begun to study the task of providing interpretable post-hoc explanations for classifier predictions. 
Such methods typically involve fitting or training a secondary interpretable mechanism on top of the base `black-box' classifier predictions. In contrast, our work directly computes importance maps from the model of interest without another layer of training (which could obfuscate the analysis).

\section{Approach}
\label{sec:approach}

\begin{figure*}[ht]
%\vskip 0.2in
\centering
%\vspace{-12pt}
\includegraphics[width=\textwidth]{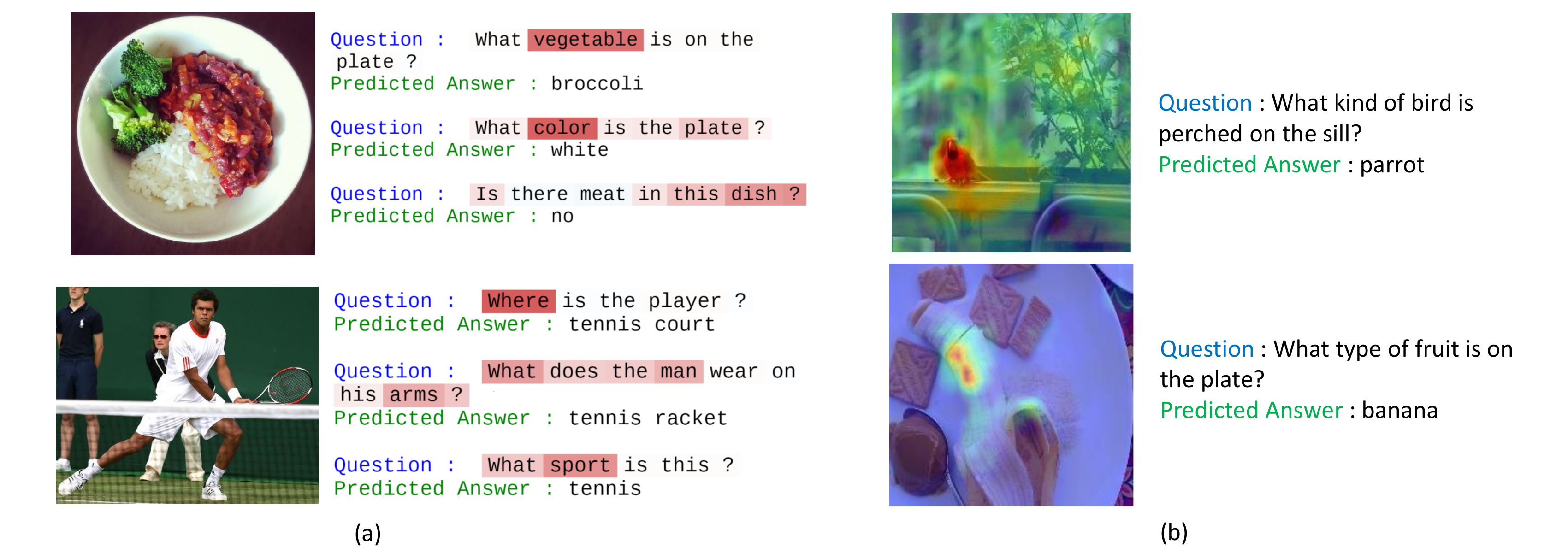}
%\vspace{-18pt}
\caption{Results for Discrete Derivatives experiment. (a) shows heat maps for questions showing the importance of words in the questions. Encouragingly, ``vegetable'' is the most important word in the first question for the predicted answer ``broccoli''.
(b) shows the importance of different regions in images. In the top image, the region containing the parrot affects the model's prediction the most.
Both visualization provide justifications for the predictions, resulting in increased transparency in the inner working of the VQA model. Best viewed in color.}
\label{fig:masked_example}

%\vskip -0.08in
\end{figure*} 

At a high level, we view a VQA model as a learned function $a = f_w(i,q)$ that takes in an input image $i$ and a question about the image $q$, is parameterized by parameters $w$, and produces an answer $a$. In order to  gauge the importance of components of $i$ and $q$ (\ie pixels and words), we consider the \emph{best linear approximation} to $f$ around each test point $(i_{test}, q_{test})$:
%\vspace{-5pt}
\begin{equation}
\begin{split}
f(i,q) \underbrace{\simeq}_{\text{best linear fit}} f(i_{test}, q_{test})  \qquad \qquad \qquad  \\
 + [i-i_{test}, q - q_{test}]^{\intercal} \nabla f(i_{test}, q_{test}) &
\end{split}
\end{equation}
% \vspace{-2pt}
Intuitively, the two key quantities we need to compute are
$\del f(i_{test}, q_{test}) / \del i$ 
and
$\del f(i_{test}, q_{test}) / \del q$, 
\ie the partial derivatives of the function \wrt each of the inputs (image and question). These expressions superficially look similar to gradients computed in backpropagation-based training of neural networks. However, there are two key differences -- (i) we compute partial derivatives of the probability of \emph{predicted output}, not the ground-truth output; and (ii) we compute partial derivatives \wrt \emph{inputs} (\ie image pixel intensities and word embeddings), not parameters. 

Due to linearization above, elements of these partial derivatives tell us the effect of those pixels/words on the final prediction. These may be computed in the following two ways.

\subsection{Guided Backpropagation}
\label{sec:guidedback}
 
Guided backpropagation \cite{GuidedBackprop} is a gradient-based visualization technique used to visualize activations of neurons in different layers in CNNs. 
It has been shown to perform better than its counterparts such as deconvolution \cite{Zeiler2014} especially for visualizing higher order layers.
Intuitively speaking, it is a modified version of backpropagation that restricts negative gradients from flowing backwards towards input layer, resulting in sharper image visualizations.

Specifically, Guided BP is identical to classical BP except in the way the backward pass is computed in Rectified Linear Units (ReLUs). Let $h^{l}$ denote the input to layer $l$ and $h^{l+1}$ denote the output. Recall that a ReLU is defined as $h^{l+1} = relu(h^{l}) = \max(h^l, 0)$. 
Let $G^{l+1} = \nicefrac{\del f}{\del h^{l+1}}$ 
% Let $G^{l+1} = \del f / \del h^{l+1}$ 
denote the partial derivative \wrt the output of the ReLU (received as input in the backward pass). The key difference between the two backprops (BP) is:
%
%\vspace{-11pt}
% \begin{align}
\begin{equation}
G^{l} = \ind{h^l > 0} \cdot G^{l+1}  \qquad \qquad \text{ [Classical BP]}
\end{equation}
\begin{equation}
G^{l} = \ind{h^l > 0} \cdot \ind{G^{l+1} > 0} \cdot G^{l+1}  \text{[Guided BP]}
\end{equation}
%\vspace{-4pt}
% \end{align}
\ie guided BP blocks negative gradients from flowing back in ReLUs. 
For more details, please refer to \cite{GuidedBackprop}.

% \change{In this method, we apply guided backpropagation to compute gradients\footnote{Note that these computations are technically not gradients since they are biased for interpretability of visualizations.} of the probability of the predicted answer with respect to the inputs-- question and image.
% The language part of the model does not contain any ReLU units, hence the gradients received using guided backpropagation are, in fact, exact gradients. 
% }
% We consider the words in the question / pixels in the image with the highest gradients as most important for the model while predicting the answer, because changing these inputs will affect the model's confidence the most.
We use guided BP to compute `gradients' of the probability of predicted answer \wrt inputs (image and question). Note that the language pathway in the models we typically use, does not contain ReLUs, thus these are true gradients (not just gradient-based visualizations) on the language side. We interpret the words/pixels with the highest (magnitude) gradients received as the most important for the model since small changes in these lead to largest changes in the model's confidence in the predicted answer.

% \vspace{-3pt}
\subsection{Discrete Derivatives}
\label{sec:occlusion}

In this method, we systematically occlude subsets of the input, forward propagate the masked input through the VQA model, and compute the change in the probability of the answer predicted with the unmasked original input. 
Since there are 2 inputs to the model, we focus on one input at a time, keeping the other input fixed (mimicing partial derivatives).
Specifically, to compute importance of a question word, we mask that word by dropping it from the question, and feed the masked question with original image as inputs to the model.
%and compute the difference in the probability of the original predicted answer.
The importance score of the question word is computed as the change in probability of the original predicted answer. 
% Note that this computation can sometimes result in negative importance scores in the case when masking a question word actually increases the probability of the predicted answer.

We follow the same procedure on the images to compute importance of image regions. We divide the image into a grid of size 16 x 16, occlude one cell at a time with a gray patch\footnote{a gray patch of intensities (R, G, B) = (123.68, 116.779, 103.939), mean RGB pixel values across a large image dataset ImageNet \cite{ImageNet} on which the CNN is trained.}, feed in the perturbed image with the entire question to the model, and compute the decrease in the probability of the original predicted answer.
The generated importance maps are shown in \figref{fig:masked_example}. 
% \change{These importance maps provide justifications for the predictions, resulting in increased transparency in the inner working of the VQA model.
% }

More results and interactive visualizations can be found on authors' webpages.\footnote{Question importance maps: \url{https://mlp.ece.vt.edu/masked_ques_vis/}. Image importance maps: \url{https://mlp.ece.vt.edu/masked_image_vis/}}

% \change{
% More results and interactive visualizations can be found at \url{https://mlp.ece.vt.edu/masked_ques_vis/} and \url{https://mlp.ece.vt.edu/masked_image_vis/}.
% If the mouse is hovered over a question word / region in the image, the predicted answer with that question word /image region occluded can be seen.}

\section{Results}
\label{sec:results}

While image/question importance maps on individual inputs provide crucial insight into the inner-workings of a model (\eg, see \figref{fig:masked_example}), what do the aggregate statistics of these maps tell us about the model?

\subsection{Analyzing Image Importance} 
\cite{human_attention} recently collected human attention annotations for (question, image) pairs from VQA dataset \cite{VQA}.
Given a blurry image and a question, humans were asked to deblur the regions in the image that were helpful in answering the question.

%\vspace{-5pt}
\begin{table}[h]
\setlength{\tabcolsep}{10pt}
{\small
\begin{center}
\begin{tabular}{@{}lcc@{}}
\toprule
  & Rank-correlation \\
\midrule
  Random &  0.000 $\pm$ 0.001\\
  Occlusion &  0.173 $\pm$ 0.004\\  
  Guided backpropagation & 0.292 $\pm$ 0.004\\
  Human & 0.623 $\pm$ 0.003\\
\bottomrule
\end{tabular}
\end{center}
}
% \vspace{-14pt}
\caption {Rank-correlation of importance maps with human attention maps (higher is better). The last row represents inter-human agreement.}
\label{tab:image_maps}
\end{table}
%\vspace{-8pt}

We evaluate the quality of image importance maps obtained from the two methods (guided backpropagation and occlusion) by comparing them to the human attention maps. 
The human attention dataset contains annotations for 1374 (question, image) pairs from VQA 
\cite{VQA} 
validation set.
Following the evaluation protocol in \cite{human_attention},
we take the absolute value of the importance maps and compute their mean rank-correlation with the human attention maps.
Specifically, we first scale both the image importance and human attention
maps to 14x14, normalize them spatially and
rank the pixels according to their spatial attention,
and then compute correlation between these two
ranked lists.
The results are shown in \tableref{tab:image_maps}.
We find that both importance maps (occlusion and guided BP) are weakly positively correlated with human attention maps, although it is far from inter-human correlation. 
Thus, our techniques revealed an interesting finding -- that even without attention mechanisms, VQA models may be implicitly attending to relevant regions in the image.

\subsection{Analyzing Question Importance} 

\begin{figure}[ht]
%\vskip 0.2in
\centering
%\vspace{-12pt}
\includegraphics[width=\columnwidth]{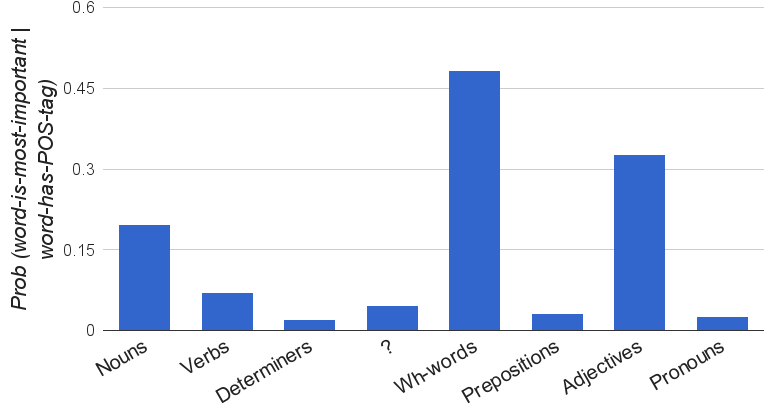}
% \vspace{-3pt}
\caption{Probability of a word being most important in a question given that it has a certain POS tag. POS tags are sorted in the decreasing order of their frequency in the entire dataset. 
%\todo{update the figure with guided backprop results}
}
\label{fig:ques_imp_hist}

% \vskip -0.1in
\end{figure} 

Since there is no human attention dataset for questions, we instead analyze the importance maps for questions using their POS tags.
Our hypothesis is that wh-words and nouns should matter most to a `sensible' model's prediction. 
We plot the probability of a word being most important in a question given that it has a certain POS tag. 
To get reliable statistics, we picked 15 most frequent POS tags from the VQA validation dataset, and grouped similar tags into one category, \eg WDT, WP, WRB are grouped as wh-words.
The histogram can be seen in \figref{fig:ques_imp_hist}.
Indeed, wh-words are most important followed by adjectives and nouns.
Adjectives and nouns rank high because many questions tend to ask about characteristics of objects, or objects themselves.
This finding suggests that the language model part of the VQA model is strong and is able to learn to focus on appropriate words without any explicit attention procedure. 
%The finding is supported by the fact that language-only version of this VQA model performs significantly better than the image-only version (VQA accuracy of 48.76 vs 28.13 \cite{VQA}).

Note that for many occlusions, the model's predicted answer is different from the original predicted answer.
In fact, we found that the number of times the predicted answer changes
correlates with the model's accuracy. It is able to predict success/failure accurately 72\% of the times.
This suggests that features that characterize these importance maps can provide useful signals for predicting the model's oncoming failures.

\section{Conclusion} 
\label{sec:conclusion}

In this paper, we experimented with two visualization methods -- guided backpropagation and occlusion -- to interpret deep learning models for the task of Visual Question Answering. 
% The first method uses guided backpropagation to analyze important words in the question, and important regions in the image. 
% In the second method, we occlude portions of the input and observe the change in prediction probabilities of the model, to compute importance of question words and image regions.
Although we focus on only one VQA model in this work, the methods are generalizable to all other end-to-end VQA models.
The occlusion method can even be applied to any (non-end-to-end) VQA model considering it as a black box.
We believe that these methods and results can be helpful in interpreting the current VQA models, and designing the next generation of VQA models.

%We further show that the computed importance maps for question words and image regions contain cues which are helpful for predicting the VQA model's likelihood of producing the correct answer.

\textbf{Acknowledgements.} 
%We thank blah blah for helpful suggestions and discussions. 
This work was supported in part by the following: 
National Science Foundation CAREER awards to DB and DP, 
Army Research Office YIP awards to DB and DP, 
ICTAS Junior Faculty awards to DB and DP, 
Army Research Lab grant W911NF-15-2-0080 to DP and DB, 
Office of Naval Research grant N00014-14-1-0679 to DB, 
Paul G. Allen Family Foundation Allen Distinguished Investigator award to DP, 
Google Faculty Research award to DP and DB,
AWS in Education Research grant to DB, and NVIDIA GPU donation to DB. 
The views and conclusions contained herein are those of the authors and should not be interpreted as necessarily representing the official policies or endorsements, either expressed or implied, of the U.S. Government or any sponsor.

\bibliography{emnlp2016}
\bibliographystyle{emnlp2016}

\end{document}